# Identification of Capture Phases in Nanopore Protein Sequencing Data Using a Deep Learning Model

Annabelle Martin[1], Daphne Kontogiorgos-Heintz[1], Jeff Nivala[1]

[1]University of Washington

**Abstract**

Nanopore protein sequencing produces long, noisy ionic current traces in which key molecular phases, such as protein capture and translocation, are embedded. Capture phases mark the successful entry of a protein into the pore and serve as both a checkpoint and a signal that a channel merits further analysis. However, manual identification of capture phases is time-intensive, often requiring several days for expert reviewers to annotate the data due to the need for domain-specific interpretation of complex signal patterns. To address this, a lightweight one-dimensional convolutional neural network (1D CNN) was developed and trained to detect capture phases in down-sampled signal windows. Evaluated against CNN-LSTM (Long Short-Term Memory) hybrids, histogram-based classifiers, and other CNN variants using run-level data splits, our best model, CaptureNet-Deep, achieved an F1 score of 0.94 and precision of 93.39% on held-out test data. The model supports low-latency inference and is integrated into a dashboard for Oxford Nanopore experiments, reducing the total analysis time from several days to under thirty minutes. These results show that efficient, real-time capture detection is possible using simple, interpretable architectures and suggest a broader role for lightweight ML models in sequencing workflows.

**Introduction**

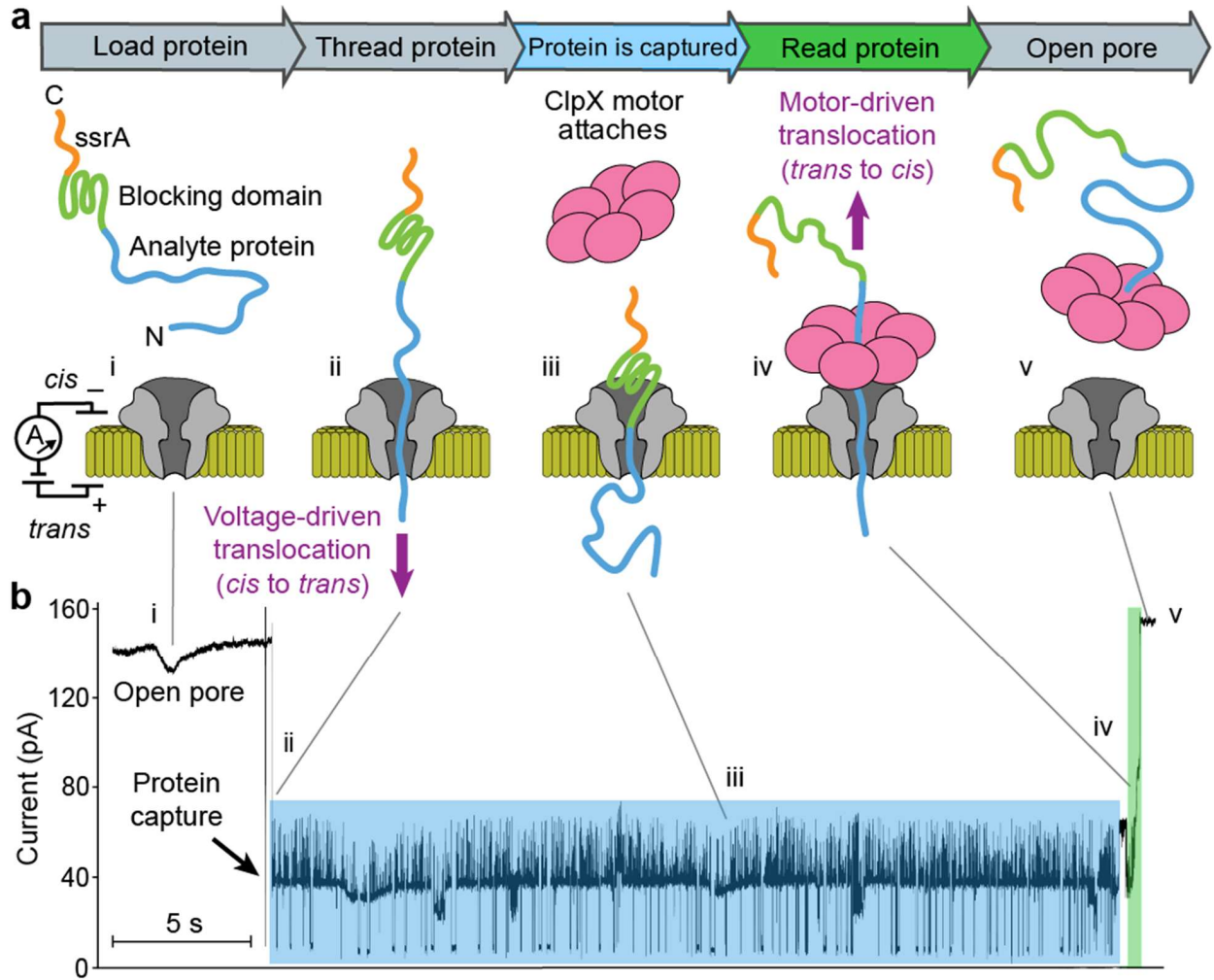

**Figure 1. Overview of protein sequencing using a nanopore system.** *(a)* Schematic of the sequencing process: proteins are threaded into the nanopore. Capture occurs when the protein enters the pore and interrupts ionic current but is not yet translocated. Proteins are then translocated by the ClpX motor. *(b)* Representative ionic current trace during a sequencing run, with key phases labeled (i through v). The noisy sustained signal with drops to near zero are capture sections (iii).

**Nanopore Protein Sequencing**

Nanopore sequencing is a powerful platform for single-molecule analysis of nucleic acids and proteins. This technique was first attempted by the Molecular Information Systems Lab (MISL) in 2013, and has been developed over the past decade (Nivala et al., 2013). By monitoring ionic current through a nanoscale pore, this technology detects changes in current as molecules translocate through the pore. Recent advances have extended this approach to full-length protein sequencing. In Oxford Nanopore's systems, such as the MinION, a ClpX unfoldase motor threads proteins into a CsgG pore in a stepwise fashion. The result is a high-resolution, continuous time-series of ionic current measurements that reflect the state and movement of the molecule. Distinct phases of this signal are characteristic of key phases (Figure 1): the open-pore phase exhibits a high baseline current (~180 pA), the closed-pore state shows suppressed current (~5 pA), the capture phase presents a noisy, sustained low current (~20 pA) often with intermittent drops, and the translocation phase features ClpX-driven, step-like changes as the protein passes through.

**Why Capture Phases Matter**

Before translocation can begin, the protein must first enter the pore, which is called a capture phase. This phase is meaningful as it indicates successful engagement of the protein with the pore complex and is experimentally crucial for monitoring channel activity via detecting captured proteins and ensuring useful signal collection. Captures precede translocations, and they can differ across proteins, experimental conditions, or flow cells. Manual labeling is not only time-consuming but also subject to user interpretation and inconsistency, making it an impractical approach for large-scale or high-throughput experiments. While machine learning has been applied to nanopore data in various contexts, including DNA sequencing and protein identification, prior studies have not focused on capture phase detection in protein nanopore sequencing. To our knowledge, this is the first work to develop and deploy a real-time machine learning model specifically trained to detect capture phases in protein nanopore current traces.

**Problem Statement**

Given the increasing use of nanopore sequencing in high-throughput settings, there is a pressing need for automated tools that can identify and interpret the distinct phases of the current signal. In particular, accurate detection of capture phases could enable real-time decision-making during experiments, improve the resolution of downstream analyses, and facilitate the creation of cleaner, more informative datasets. However, developing such a tool is non-trivial due to the noisy and diverse nature of raw signal traces, the variability in capture phase signatures, and the imbalance between capture and non-capture regions in training data.

**Objective and Approach**

The objective is to develop a method that is both accurate and computationally efficient, enabling its integration into live experimental workflows. By leveraging characteristic patterns in ionic current signals, the model automates the identification and boundary annotation of capture phases, which are crucial steps for downstream data analysis and quality control.

To achieve this, we trained a one-dimensional convolutional neural network (1D CNN) on manually annotated datasets collected in the Molecular Information Systems Lab (MISL) at the University of Washington (Motone et al., 2024). Raw current traces were down-sampled and segmented into fixed-length windows, each manually labeled as capture or non-capture based on signal morphology. The model outputs per-window confidence scores that are post-processed into binary predictions and aggregated into capture phase boundaries. Multiple model architectures, including CNN-LSTM hybrids and histogram-based classifiers, were evaluated using rigorous run-level data splits to ensure generalizability. The best-performing model was deployed in a custom PyQt5-based dashboard, enabling low-latency inference and real-time visualization across 512 sequencing channels. This pipeline forms the foundation for robust, real-time capture detection in large-scale nanopore sequencing workflows.

**Methods**

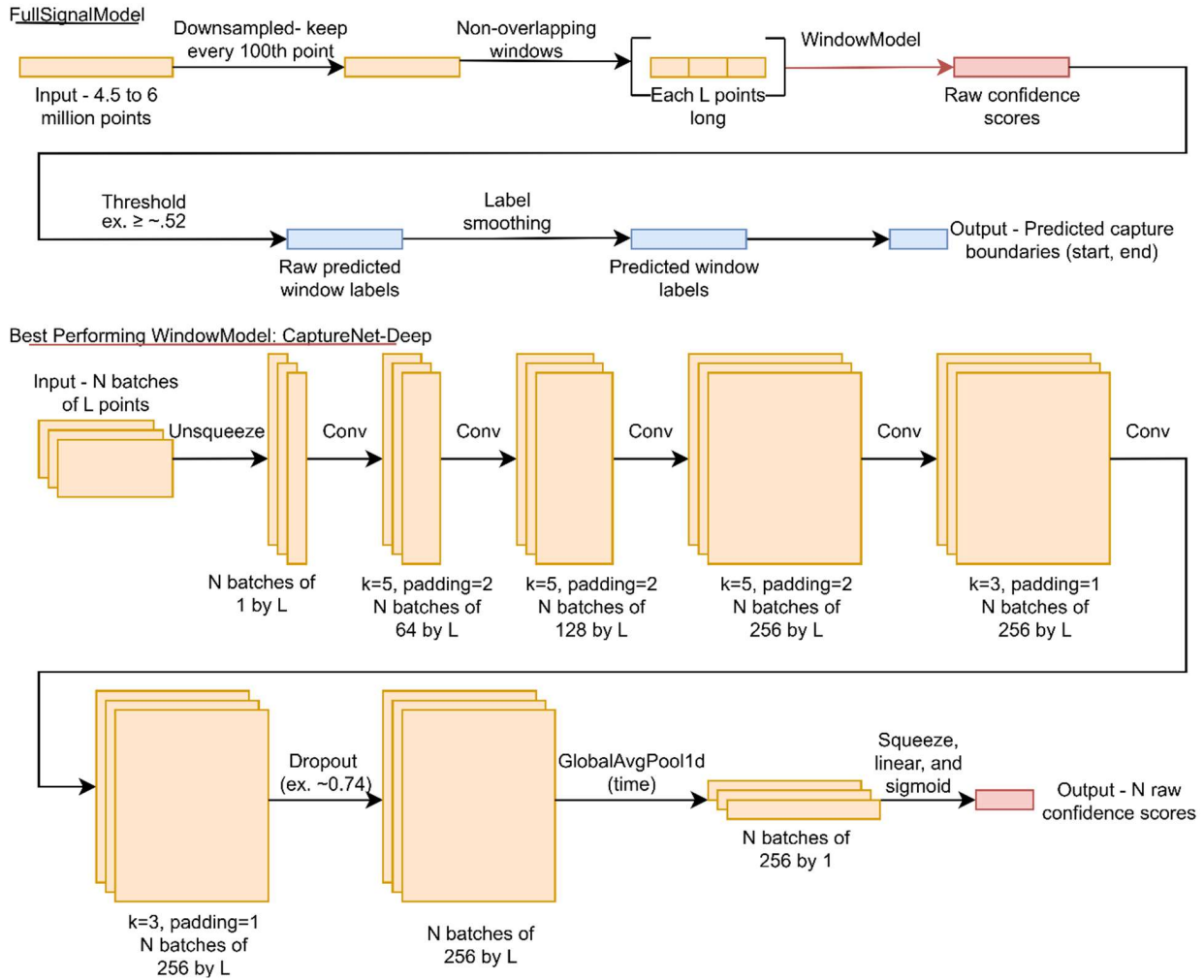

{width="6.4875in" height="4.664583333333334in"}

**Figure 2: Architecture of the FullSignalModel.** The FullSignalModel processes long current sequences (4.5--6 million data points) by first down-sampling and dividing them into windows (typically 2,000 points each for the best model), with a step size of 1.1 times the window size for training, otherwise a step size is the same as the window size. Each window is passed into the WindowModel: a convolutional neural network that outputs a confidence score indicating the likelihood that the window contains or is part of a capture phase. These confidence scores are put

through a threshold and smoothed to generate final binary labels for each window, which are then aggregated to produce the predicted capture phase boundaries.

**Data Collection and Preprocessing**

Nanopore protein sequence data was collected from 17 unique experimental runs on Oxford Nanopore's sequencing device. A significant portion of the data was sourced from signals with known translocation events, as identified by an existing translocation finder or manual labeling. Since every translocation is preceded by a capture phase, these regions provided a reliable foundation for locating and labeling capture sections. Each signal consisted of 4.5 to 6 million time-series data points representing ionic current measurements. Signals were manually labeled using visual inspection and domain expertise using LabelStudio (Tkachenko et al., 2024), identifying the boundaries of capture phases, which is a characteristic phase marked by a sustained, noisy current of approximately 20 pA (picoamps), accompanied by brief, sharp drops to near zero pA. Signals can have 0 or more capture phases. These annotated sequences were used to train the FullSignalModel, which takes one of the sequences as the input and generates the predicted boundaries of the capture phases.

**Model Architecture**

The FullSignalModel takes an input of the full 4.5-6 million point long sequence. To reduce computational load and improve model efficiency for real-time use, each sequence is downsampled by a factor of 100. This maintains enough resolution for the model to detect capture phase signatures but makes the model significantly more computationally efficient. These data sequences are divided into non-overlapping, fixed-length windows with a step size of 1.1 times the window length. The optimal window size was found to be 2,000 points for the best-performing model. To accurately detect capture phases from these windows, a hybrid neural network architecture was developed, called the WindowModel. Multiple WindowModel architectures were developed and evaluated.

The best-performing architecture, CaptureNet-Deep, employs a more robust convolutional neural network design with additional regularization. This model consists of multiple 1D CNN layers with dropout regularization to prevent overfitting. The 1D CNN component was designed to extract local temporal features from each windowed current trace. These features capture subtle shape patterns in the signal, including sharp drops, low sustained currents, and noise characteristics associated with capture behavior.

Each convolutional layer uses appropriate kernel sizes and padding to preserve important temporal relationships while extracting hierarchical features. After the convolutional layers, global average pooling is applied across the sequence dimension to aggregate temporal information into a single feature vector per sample. This pooled vector is then passed through fully connected layers with dropout (dropout rate of ~0.74) that maps to a single output value. A sigmoid activation function is applied to produce the final probability indicating the likelihood of a capture phase.

The WindowModel outputs a single scalar confidence score between 0 and 1, representing the predicted likelihood that the input window contains a capture phase. These confidence scores are thresholded (threshold of ~0.52) to produce binary labels. To reduce prediction noise and account for small misclassifications, a label smoothing heuristic was applied: isolated windows surrounded by opposing labels were flipped to match their neighbors (e.g., [0, 1, 0] → [0, 0, 0] and [1, 0, 1] → [1, 1, 1]). The final array of window-level predictions was then aggregated to determine the start and end indices of each predicted capture phase. These predicted capture boundaries constitute the output of the FullSignalModel. A diagram of these models can be seen in Figure 2.

**Training Procedure**

To train the FullSignalModel, the dataset was split into training (72%), validation (18%), and test (10%) sets at the run level to prevent data leakage and overfitting. This run-level split ensures that windows from the same experimental run do not appear in both training and test sets, providing a more realistic evaluation of model generalizability. Only runs with valid, annotated current sequences were used.

Given the severe class imbalance in the raw data with capture phases being significantly rarer than non-capture segments, a two-step strategy was implemented. First, non-capture windows (defined as segments with less than 50% overlap with any annotated capture section) were only included if they helped maintain a 1:1 ratio between capture and non-capture windows. This approach preserved a balanced training set without requiring synthetic sampling. Second, the model was trained using the Binary Cross-Entropy (BCE) loss function, which is well-suited for binary classification problems.

All models were trained using automated hyperparameter optimization implemented with the Optuna framework. For each model architecture, 40 optimization trials were conducted to systematically search the hyperparameter space and identify optimal configurations. The optimization process used a Bayesian approach to intelligently sample hyperparameter combinations based on previous trial results.

The hyperparameter search spaces varied by model but typically included: window size (1000-3000 points), learning rate (1e-5 to 1e-2, log-uniform), weight decay (1e-6 to 1e-2, log-uniform), batch size (32-1024 depending on model complexity), dropout rate (0.0-0.8 for applicable models), and decision threshold (0.3-0.7). Additional architecture-specific parameters were optimized, such as LSTM hidden size and number of layers for the CNN-LSTM hybrid, number of histogram bins for histogram-based models, and FPN channel dimensions for pyramid network variants.

The optimization objective was a weighted combination of classification metrics: maximize_score = (1×Accuracy + 3×Precision + 1×Recall + 2×F1) / 7, which prioritizes precision while maintaining balanced performance across all metrics. Precision was assigned a higher weight in the maximize_score function to reflect the priority of minimizing false positives, which is critical in identifying high-quality, clean data. The selection of weights was guided by domain-specific requirements rather than empirical optimization; alternative weighting schemes were not explored in this study. Training was performed over a maximum of 500 epochs with early stopping (patience of 50 epochs) monitoring validation loss.

The best-performing WindowModel was identified to be a CaptureNet-Deep model architecture with a window size of 2,000 points, step size of 2,200, batch size of 128, learning rate of 0.000183, weight decay of 0.00332, dropout rate of 0.739, and decision threshold of 0.524. All models used the AdamW optimizer, which was found to provide superior convergence compared to standard Adam in preliminary experiments. All experiments were conducted using PyTorch on GPU hardware with CUDA acceleration. Data processing and training pipelines utilized standard Python libraries, including NumPy, pandas, and scikit-learn.

**Real-Time Deployment**

To enable real-time monitoring and feedback during protein sequencing experiments, a custom visualization dashboard was designed and developed for integration with the trained FullSignalModel. This desktop application, developed using PyQt5 and PyQtGraph, supports offline analysis of recorded .fast5 files. A proof-of-concept version of the dashboard with real-time streaming from Oxford Nanopore MinKNOW instruments was successfully demonstrated.

The dashboard displays all 512 channels as a grid of miniature plots, each dynamically color-coded based on signal activity: blue for detected capture phases, dark red for dead pores (low signal activity), and light green for translocation events. Any channel can be clicked to open a large, detailed view showing the raw signal over time, overlaid with predicted capture and translocation event regions. Users can analyze recorded .fast5 files through the dashboard's file import mode, which supports both local and SSH-based remote browsing. The user interface includes tools for visual customization (e.g., Y-axis range control), and export functionality. Detected capture regions, confidence values, translocations, and pore status (e.g., flagged as dead) can be exported as structured .json files for downstream analysis.

During live experiments, the dashboard connects to MinKNOW via the minknow_api interface and streams ionic current signals from all 512 channels. Signals are periodically down-sampled and partitioned into windows using the same parameters as the trained model. Each window is passed through the WindowModel which estimates the likelihood that the window corresponds to a capture phase. Confidence scores above a threshold are binarized and smoothed using the same strategy as during training, which is then aggregated to produce contiguous predictions. The real-time dashboard has not been fully tested with all functionality.

# Results

## Evaluation Metrics

The best-performing model was CaptureNet-Deep with chosen hyperparameters as described above. This model achieved an accuracy of 93.19%, precision of 93.39%, recall of 95.39%, and an F1 score of 0.94 on the held-out test set. The model demonstrates strong performance across all metrics, with particularly high recall indicating effective recovering of true positive phases while maintaining good precision to minimize false positives.

## Model Comparison

| Model Name | Accuracy | Precision | Recall | F1 | Maximize score |
|---|---|---|---|---|---|
| **CaptureNet-Deep** | 93.19 | 93.39 | 95.39 | 0.94 | 93.84 |
| **CaptureCNNWithLSTM** | 91.75 | 94.36 | 91.78 | 0.93 | 93.11 |
| **RandomForestHistogram** | 91.95 | 92.33 | 94.34 | 0.93 | 92.85 |
| **HistogramCaptureClassifier** | 91.24 | 91.40 | 93.88 | 0.92 | 92.04 |
| **FPNCaptureDetector** | 90.82 | 92.23 | 92.20 | 0.92 | 91.97 |
| **CaptureCNNnet** | 90.73 | 91.37 | 93.17 | 0.92 | 91.73 |
| **SpikeClassifier** | 82.16 | 78.07 | 96.92 | 0.86 | 83.64 |

**Table 1: Model Performance Comparison.** Table 1 summarizes the performance of different architectures evaluated on the held-out test set using run-level data splits. Each model was optimized using automated hyperparameter search, and only the best-performing configuration per model type is shown. All models were trained and evaluated using proper run-level splits to prevent overfitting.

CaptureNet-Deep achieved the highest F1 and recall, confirming that added architectural complexity improved sensitivity to capture events while maintaining good precision. Interestingly, the CNN-LSTM hybrid (CaptureCNNWithLSTM) achieved the highest precision (94.36%) but lower recall, suggesting that while it is conservative in its predictions, it may miss some subtle capture phases. The histogram-based RandomForestHistogram performed surprisingly well, indicating that statistical features of the signal contain significant discriminative information. However, the pure CNN approaches generally outperformed histogram-based methods, suggesting that local temporal patterns captured by convolutions provide additional valuable information for capture detection.

## Visualization of Predictions

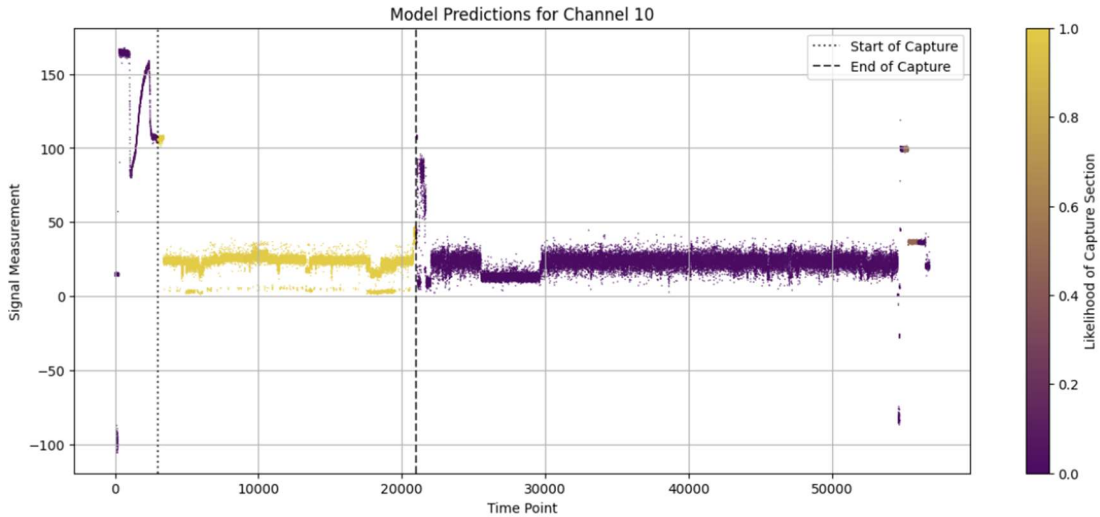

{width="6.5in" height="2.9208333333333334in"}

**Figure 3: Model predictions for Sample A.** A heatmap-style dot plot showing model-generated capture likelihoods across a 60,000-point current trace (down-sampled by a factor of 100 from 6,000,000). Color represents the predicted probability that each time point belongs to a capture section (Purple = low likelihood, gold = high likelihood). The black dashed lines indicate the ground truth capture section boundaries manually labeled for this sequence.

In Figure 3, the model successfully identifies a single, sustained region of high capture likelihood between the manually annotated ground truth boundaries. The confidence remains consistently high throughout the labeled region, while flanking regions remain confidently non-capture. This prediction demonstrates the model's ability to sharply distinguish between capture and non-capture behavior, particularly in sequences with a clearly defined capture phase and minimal noise.

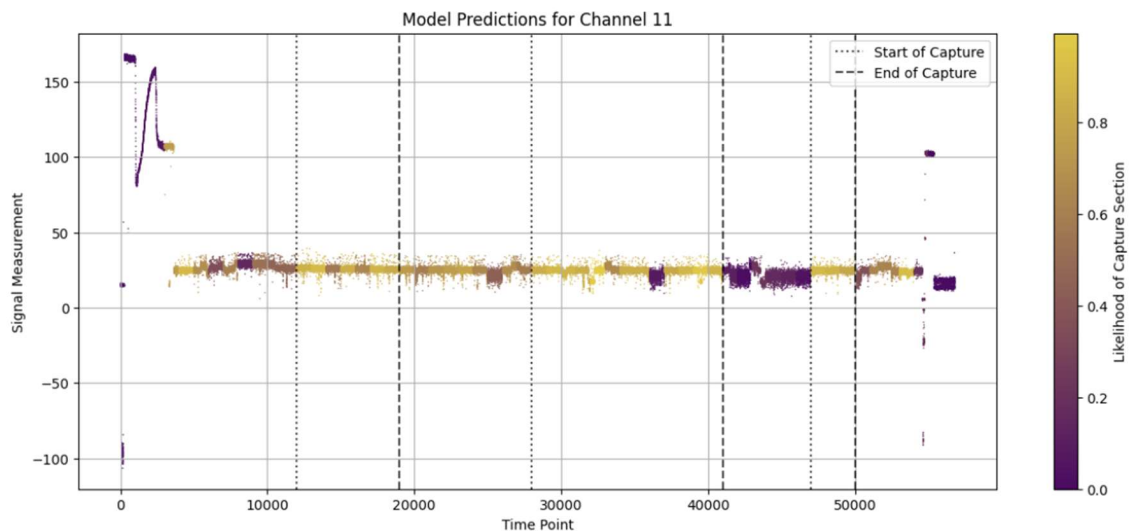

{width="6.5in" height="2.9055555555555554in"}

**Figure 4: Model predictions for Sample B.** This figure visualizes model predictions for a more complex input, which includes three capture sections. The heatmap coloring follows the same scale as in Figure 3. Multiple black dashed lines show the labeled capture boundaries.

In Figure 4, the model exhibits both false negatives and false positives. Some windows in the capture phase are either missed or predicted with very low confidence, as indicated by the purple color. Additionally, the model incorrectly predicts noisy regions in the sequence as capture phases. These errors suggest that while the model is capable of identifying capture phases, its precision may degrade when capture phases are shorter, less distinct, or embedded in high-noise regions.

**Oxford Nanopore Data Visualization Dashboard**

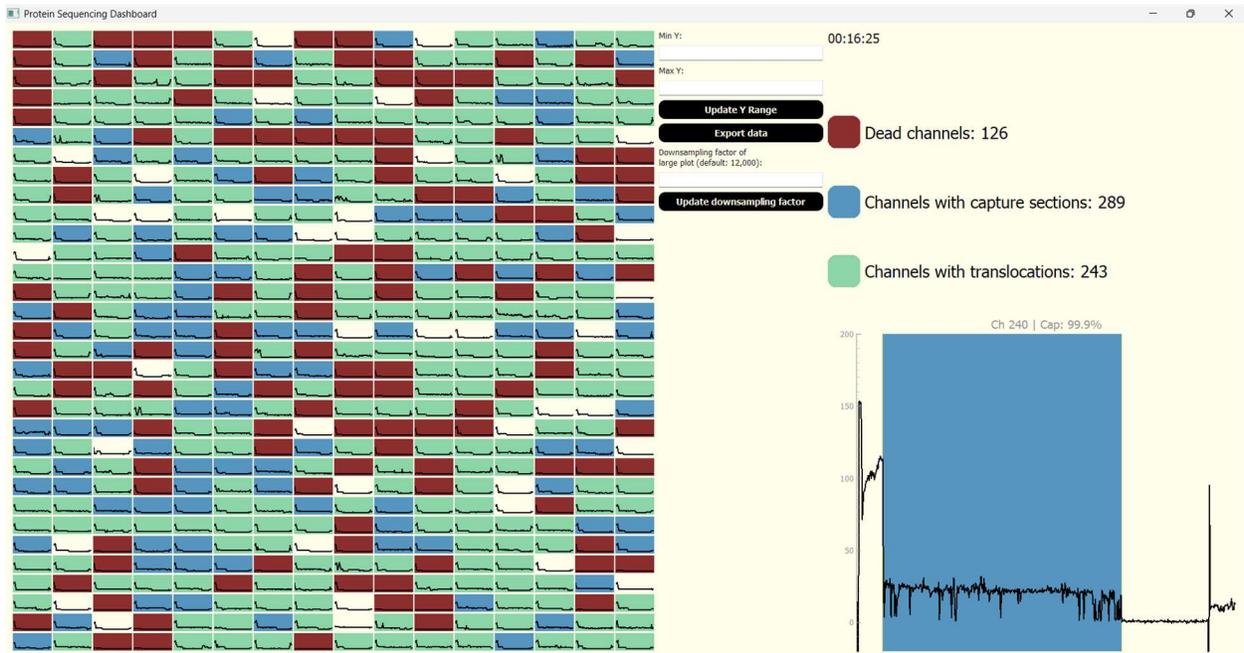

{width="6.5in" height="3.813888888888889in"}

**Figure 5: Interface of the Dashboard.** A live data dashboard developed to visualize and analyze nanopore protein sequencing signals in real time. Each small plot on the left represents a single channel from the sequencing device. Channels containing model-detected capture phases are highlighted in blue, while dead or inactive channels are marked in red. The main signal for a selected channel (bottom right) is displayed in detail with shaded blue regions indicating predicted capture sections and the associated model confidence score. The panel also summarizes the number of dead channels, capture-active channels, and translocation-active channels (as detected by a separate model not described in this paper).

The dashboard provides real-time visibility into sequencing activity across all channels and integrates the capture detection model for near-live inference. Predicted capture phases are automatically highlighted on the signal trace of selected channels, enabling fast inspection of candidate regions during experimental runs. The CaptureNet-Deep model maintains efficient inference performance suitable for real-time deployment, processing signal data with minimal latency. Within the multithreaded dashboard pipeline, which includes I/O and process management overhead, the system provides responsive analysis of incoming data streams. Instead of having expert reviewers review the data over several days, the dashboard can analyze the data in less than 30 minutes, leading to faster experimental turn-around rate.

## Discussion/Conclusion

### Summary of Findings

This work presents a convolutional neural network model for detecting capture phases in nanopore protein sequencing data. By processing down-sampled current traces in fixed-length windows, the model learns to distinguish capture regions based on local signal features. Among several architectures evaluated using proper run-level data splits, CaptureNet-Deep achieved the best overall performance, with an F1 score of 0.94, precision of 93.39%, and recall of 95.39% on the test set. The model's performance demonstrates that automated capture detection is feasible for real-time experimental monitoring and downstream analysis workflows.

### Interpretation and Implications

The strong performance of the enhanced CNN architecture (CaptureNet-Deep) suggests that capture phases are characterized by local signal patterns that can be effectively identified through hierarchical feature extraction with appropriate regularization. The improved performance over simpler CNN variants indicates that additional model capacity and dropout regularization help generalize to unseen experimental runs. The fact that CNN-based approaches generally outperformed histogram-based methods suggests that temporal pattern recognition provides advantages over purely statistical approaches, though both of these capture complementary aspects of the signal.

The use of proper run-level data splits revealed more realistic performance estimates compared to window-level splits, highlighting the importance of rigorous evaluation methodology in time-series classification tasks. The resulting performance metrics provide a more reliable assessment of the model's ability to generalize to new experimental conditions.

### Limitations

Despite its strong performance, the model has several limitations. The model's generalizability to datasets from different experimental contexts, protein types, or sequencing conditions remains to be fully validated. While the model performs well on typical capture phases, it may struggle with edge cases such as very short, highly noisy, or ambiguous phases. The training labels were created through manual visual inspection, introducing potential subjectivity and limiting scalability to very large datasets. The model parameters were optimized for the current dataset and may require retuning for different experimental setups or signal characteristics. Notably, the annotation process lacked assessment of inter-annotator agreement, which raises concerns about labeling consistency. Future work would benefit from larger, more diverse datasets and multiple expert annotators to enhance model robustness and reliability.

**Future Work**

To improve generalizability, future work should evaluate the model on datasets from more diverse experimental conditions, protein types, and sequencing platforms. Incorporating a broader range of data could enhance accuracy and help mitigate labeling subjectivity. At present, the accuracy of the manual annotations likely imposes an upper bound on model performance Exploring adaptive or multi-scale windowing techniques may yield more precise boundary detection, especially for capture phases of varying duration. A more detailed error analysis, including performance stratification by capture phase characteristics (length, amplitude, noise level), could identify systematic weaknesses and guide future improvements.

Additionally, investigating ensemble methods that combine the strengths of different architectures (e.g., CNN and histogram-based approaches) could further improve performance. Integration with other nanopore analysis tools and validation on larger, multi-laboratory datasets would strengthen the model's utility for the broader nanopore sequencing community.

**Conclusion**

This work presents an effective convolutional neural network capable of accurately detecting capture phases in nanopore protein sequencing data. Through proper experimental design, including run-level data splits, CaptureNet-Deep achieves strong performance across multiple metrics while maintaining efficiency suitable for real-time deployment. The model's integration with a dashboard demonstrates its practical utility for automated experimental workflows. These results establish a foundation for more sophisticated, automated analysis tools in nanopore sequencing and highlight the importance of rigorous evaluation methodology in developing machine learning models for time-series biological data.

**Citations**

1. Motone, K., Kontogiorgos-Heintz, D., Wee, J. *et al.* Multi-pass, single-molecule nanopore reading of long protein strands. *Nature* **633**, 662--669 (2024). https://doi.org/10.1038/s41586-024-07935-7

2. Nivala, J., Marks, D. & Akeson, M. Unfoldase-mediated protein translocation through an α-hemolysin nanopore. Nat Biotechnol 31, 247–250 (2013). https://doi.org/10.1038/nbt.2503

3. Tkachenko, M., Malyuk, M., Holmanyuk, A., & Liubimov, N. (2024). Label Studio (Version 1.12.1) [Computer software]. HumanSignal. https://github.com/HumanSignal/label-studio